\pdfoutput=1
\documentclass[10pt,a4paper]{article}

\usepackage[T1]{fontenc}
\usepackage[utf8]{inputenc}
\usepackage{lmodern}
\usepackage{mathptmx}
\usepackage[scaled=0.92]{helvet}
\usepackage{microtype}
\usepackage[margin=2.5cm]{geometry}
\usepackage{graphicx}
\usepackage{booktabs,tabularx,array}
\usepackage{amsmath}
\usepackage{xcolor}
\usepackage{url}
\usepackage[super,comma,sort&compress]{natbib}
\newcommand{\linenumbers}{}\newcommand{\nolinenumbers}{}
\usepackage{caption}
\usepackage{float}
\usepackage{placeins}
\usepackage{textcomp}
\usepackage{listings}
\usepackage[skins]{tcolorbox}
\usepackage{titlesec}
\usepackage{hyperref}


\renewcommand{\arraystretch}{1.15}
\DeclareCaptionLabelSeparator{pipe}{\ $|$\ }
\lstdefinestyle{llmprompt}{
  basicstyle=\ttfamily\footnotesize,
  columns=fullflexible,
  keepspaces=true,
  breaklines=true,
  breakatwhitespace=true,
  breakindent=0pt,
  showstringspaces=false,
  upquote=true,
  frame=single,
  framerule=0.3pt,
  rulecolor=\color{rulec},
  backgroundcolor=\color{panelbg},
  framesep=5pt,
  xleftmargin=5pt,
  xrightmargin=5pt,
  aboveskip=0.3\baselineskip,
  belowskip=0.3\baselineskip,
  literate={’}{{\textquoteright}}1 {—}{{\textemdash}}1
}

\newcommand{\cat}[1]{\texttt{#1}}
\newcommand{\file}[1]{\path{#1}}

\definecolor{accent}{HTML}{3B5B7E}
\definecolor{ink}{HTML}{22303D}
\definecolor{rulec}{HTML}{CAD4DE}
\definecolor{panelbg}{HTML}{F4F7FA}
\hypersetup{colorlinks=true, linkcolor=accent, citecolor=accent, urlcolor=accent,
  pdftitle={A dataset of clinically similar image pairs from citation-linked case reports for medical image retrieval},
  pdfauthor={Guy Shmueli, Keno K. Bressem, Tom Hope}}
\titleformat{\section}{\sffamily\large\bfseries\color{ink}}{}{0pt}{}
\titlespacing*{\section}{0pt}{2.6ex plus .6ex minus .2ex}{0.9ex}
\titleformat{\subsection}{\sffamily\normalsize\bfseries\color{ink}}{}{0pt}{}
\titlespacing*{\subsection}{0pt}{1.8ex plus .4ex minus .2ex}{0.4ex}
\newcommand{\parhead}[1]{\textbf{#1}}
\newcommand{\linkrow}[4]{\makebox[1.6em][l]{\raisebox{-0.2em}{\includegraphics[height=1.05em]{#1}}}\textbf{#2:} &
  \href{#3}{\color{ink}\texttt{#4}}\\}

\begin{document}

\begin{tcolorbox}[enhanced, colback=panelbg, colframe=panelbg, boxrule=0pt, arc=3.5mm,
  left=7mm, right=7mm, top=6mm, bottom=4.5mm, before skip=0pt, after skip=1.4em]
{\centering
{\fontsize{17.5}{21.5}\selectfont\bfseries\color{ink}A dataset of clinically similar image pairs from\\ citation-linked case reports for medical image retrieval\par}
\vspace{1em}
{\large Guy Shmueli\textsuperscript{1}, Keno K. Bressem\textsuperscript{2,3,4,5} and Tom Hope\textsuperscript{1,6,*}\par}
\vspace{0.7em}
{\small\textsuperscript{1}School of Computer Science and Engineering, The Hebrew University of Jerusalem, Jerusalem, Israel\\
\textsuperscript{2}National Center for Tumor Diseases West, Essen, Germany\\
\textsuperscript{3}Institute of Artificial Intelligence in Medicine, University Hospital Essen, Essen, Germany\\
\textsuperscript{4}Department of Diagnostic and Interventional Radiology, School of Medicine and Health,\\ Technical University of Munich, Munich, Germany\\
\textsuperscript{5}Department of Diagnostic and Interventional Radiology and Neuroradiology,\\ University Hospital Essen, Essen, Germany\\
\textsuperscript{6}Allen Institute for AI, Seattle, WA, USA\par}
\par}
\vspace{1.2em}
\textbf{Abstract.}\enspace Retrieving images of patients clinically similar to a given patient can support diagnosis, teaching and research. Training and evaluating retrieval systems requires knowing which images of different patients show the same finding, and existing datasets are small or narrow in scope. We construct PMC-PaIR, a dataset of 86,144 pairs of clinically similar images from 56,013 open-access PubMed Central case reports, which often document rare or challenging presentations. We develop an approach that follows citations, which often link case reports of similar patients. To determine what two linked cases share clinically, it prompts large language models to generate and answer hierarchical questions, from modality to diagnosis and findings, and to match images by caption using the answers. Experts find the pairs highly accurate, and this question-guided approach finds twice as many correct pairs as direct matching. Fine-tuning BiomedCLIP on PMC-PaIR improves retrieval in all modality, anatomy and diagnosis categories tested, for image-only queries and image queries with a short text naming the finding. We release the dataset and the automated pipeline, which can extend PMC-PaIR as new case reports are published.
\par\vspace{1em}
{\small\renewcommand{\arraystretch}{1.3}%
\begin{tabular}{@{}l@{\hspace{1.2em}}l@{}}
\linkrow{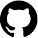}{Curation pipeline}{https://github.com/GuyShmueli/pmc-pair-pipeline}{https://github.com/GuyShmueli/pmc-pair-pipeline}
\linkrow{icon_github.pdf}{Retrieval code}{https://github.com/GuyShmueli/pmc-pair-retrieval}{https://github.com/GuyShmueli/pmc-pair-retrieval}
\linkrow{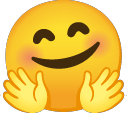}{Dataset}{https://huggingface.co/datasets/dhtandguy21/PMC-PaIR}{https://huggingface.co/datasets/dhtandguy21/PMC-PaIR}
\linkrow{emoji_hf.png}{Model}{https://huggingface.co/dhtandguy21/PMC-PaIR-BiomedCLIP}{https://huggingface.co/dhtandguy21/PMC-PaIR-BiomedCLIP}
\end{tabular}\par}
\end{tcolorbox}
{\renewcommand{\thefootnote}{\fnsymbol{footnote}}%
\footnotetext[1]{Corresponding author: Tom Hope (\href{mailto:tom.hope@mail.huji.ac.il}{tom.hope@mail.huji.ac.il}, \href{mailto:tomh@allenai.org}{tomh@allenai.org}).}}

% ===========================================================================
\section{Background \& Summary}

Clinicians facing an unfamiliar presentation, educators assembling teaching material and researchers studying a disease often need images of other patients with clinically similar findings. Content-based medical image retrieval addresses this need by ranking images by their similarity to a query image,\cite{muller2004} and retrieval from the literature can also improve the diagnostic predictions of vision--language models.\cite{mazor_lvlm_2026} Developing and evaluating such systems requires data that specify which images of different patients are clinically similar.

Published case reports could potentially be a valuable source of such images. A case report describes an individual patient in detail, including imaging, and is often published because the presentation is rare, atypical or diagnostically challenging.\cite{nissen_casereports_2014} Comparable cases are therefore scarce in any single institution's archive but can be found across the literature. The PubMed Central (PMC) Open Access Subset makes many such reports and their figures available for reuse.\cite{nlm_pmcoa_2026}

However, existing datasets of PMC figures relate each figure only to text from its own article, whether they draw on articles of any type, such as ROCOv2, MedICaT and PMC-15M,\cite{ruckert_rocov2_2024,subramanian_medicat_2020,biomedclip} or on case reports, such as MultiCaRe.\cite{nievasoffidani_multicare_2024,nievasoffidani_multicare_2025} Medical image retrieval benchmarks either cover many modalities and body regions with few queries or are restricted to a single modality or body region. ImageCLEF\cite{kalpathycramer_imageclef_2015} and the image-retrieval task of M3Retrieve\cite{acharya_m3retrieve_2025} have a few dozen to several hundred queries, whereas 3D-MIR covers computed tomography (CT) volumes of four organs\cite{benabacha_3dmir_2023} and RadIR covers chest radiographs and chest CT scans.\cite{zhang_radir_2025} None of these resources provides clinically similar image pairs at scale across modalities, body regions and diseases.

\begin{figure}[!t]
\centering
\includegraphics[width=\linewidth]{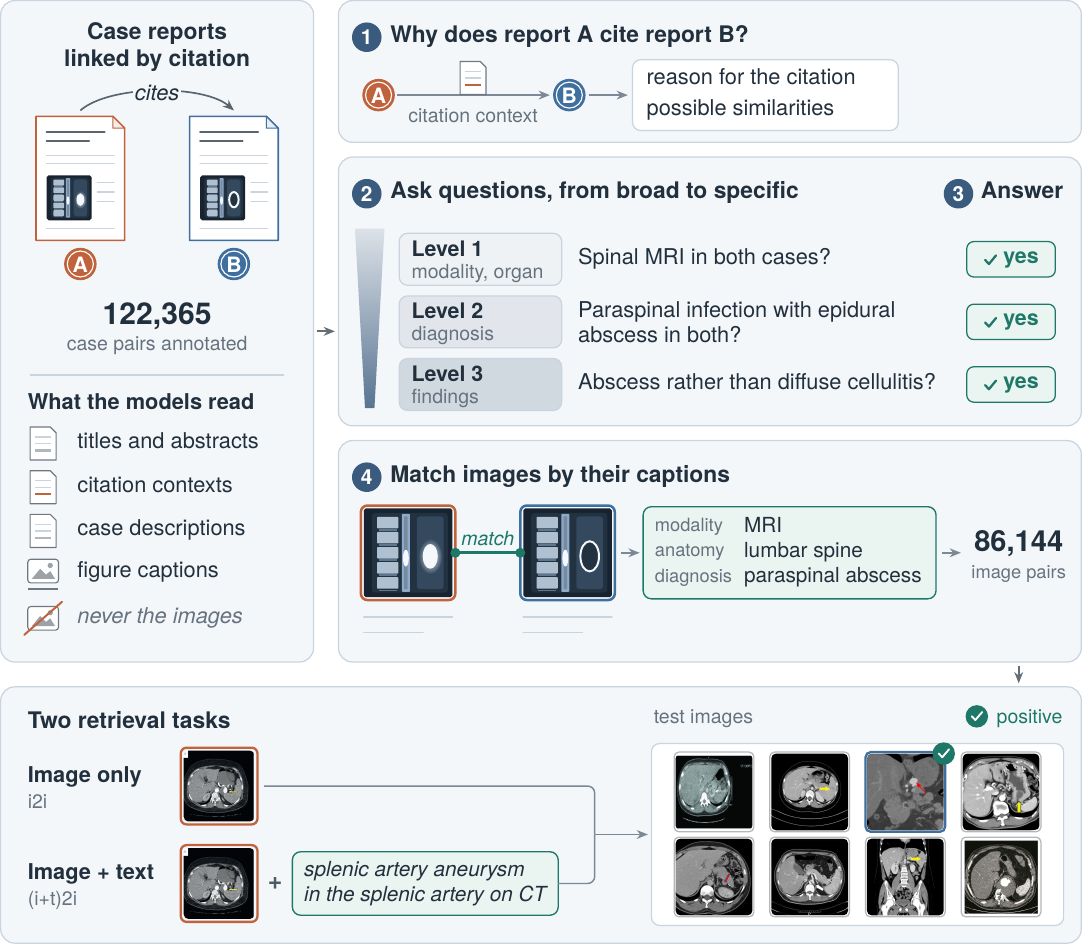}
\caption{\textbf{Construction of PMC-PaIR and the two retrieval tasks.} Cases whose source articles are linked by a citation form case pairs (left), of which 122,365 were annotated. For each case pair, language models that read only text (1)~state why report~A cites report~B, (2)~ask comparative questions that move from imaging modality and organ system to diagnosis and specific findings, (3)~answer them from the case descriptions and (4)~match images whose captions describe the same modality, anatomy and type of abnormality, recording descriptors of what they share. After post-processing, 86,144 image pairs remained (right). In the two retrieval tasks (bottom), abbreviated i2i and (i+t)2i, each test image serves as a query, either alone or with the descriptions of its pairs, and the other test images, except those linked to it only indirectly, are ranked to retrieve the images paired with it. Models are listed in Table~1, and the prompts are given at the end of Methods. The example in steps 1--4 is the case pair\cite{pmc9899453,pmc4023003} of Table~2, with abridged text and schematic images. The retrieval example is the query of Fig.~2, with images reproduced, cropped, from the source articles of the query, its positive and the other test images shown,\cite{pmc5748087,pmc8349697,pmc5949188,pmc9801666,pmc12414647,pmc6597495,pmc4980792,pmc3703373} under the licenses given in the legend of Fig.~2. CT, computed tomography; MRI, magnetic resonance imaging.}

\end{figure}
\begin{figure}[!t]
\centering
\includegraphics[width=\linewidth]{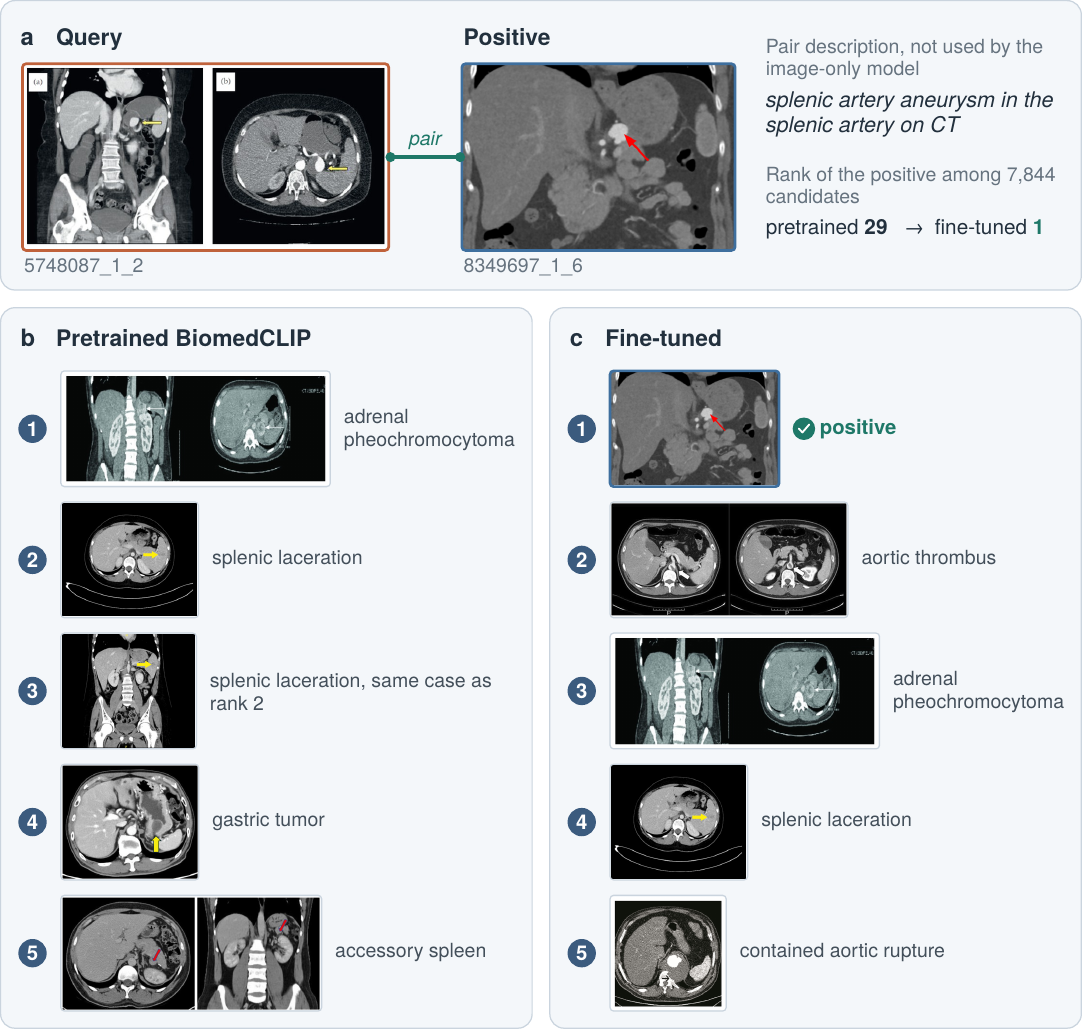}
\caption[Example of image-only retrieval]{\textbf{Example of image-only retrieval.} \textbf{a},~Test query \texttt{5748087\_1\_2}, a CT image of a splenic artery aneurysm, and its only positive, the single image paired with it in PMC-PaIR. The image-only (i2i) model does not use the pair description shown. \textbf{b},\textbf{c},~The five highest-ranked candidates of pretrained BiomedCLIP (\textbf{b}) and of the fine-tuned i2i model (\textbf{c}), labeled with the finding they show. Fine-tuning moved the positive, marked in \textbf{c}, from rank 29 to rank 1 among 7,844 candidates. The query was chosen for illustration. Images are reproduced, some cropped, from the source articles of the query,\cite{pmc5748087} the positive\cite{pmc8349697} and the candidates showing an adrenal pheochromocytoma,\cite{pmc5949188} a splenic laceration,\cite{pmc9801666} a gastric tumor,\cite{pmc12414647} an accessory spleen,\cite{pmc6597495} an aortic thrombus\cite{pmc4980792} and a contained aortic rupture,\cite{pmc3703373} under the Creative Commons Attribution 4.0 license (\url{https://creativecommons.org/licenses/by/4.0/}), except the last, which is reproduced under the Creative Commons Attribution 3.0 license (\url{https://creativecommons.org/licenses/by/3.0/}). CT, computed tomography.}

\end{figure}

We present PMC-PaIR (PubMed Central Patient Image Retrieval), a dataset of 86,144 pairs of clinically similar images from 56,013 open-access case reports (Fig.~1). Two radiological, nuclear-medicine or ophthalmic images from different reports are clinically similar when they share an imaging modality, an anatomical region and a type of abnormality, such as an abscess or an aneurysm. We build on PMC-Patients, which links the patients of two case reports whenever one cites the other.\cite{pmcpatients,pmcpatients_data} Because authors often cite earlier reports of similar patients, we form image pairs between linked reports. PMC-Patients does not, however, examine why the citation was made or link the figures of the two reports. A citation alone does not indicate which figures correspond, because a report's figures may show the primary disease, a complication, a coexisting condition or the result of treatment, and authors cite earlier reports for many reasons. Expert annotation of image pairs is costly,\cite{karimi_noisy_2020} and labels extracted automatically from radiology reports\cite{irvin_chexpert_2019} can differ from what individual images show.\cite{jain_visualchexbert_2021} For each pair of linked reports, or case pair, large language models (LLMs) therefore identify the reason for the citation from its citation contexts, compare the two cases through structured questions that progress from imaging modality and organ system to diagnosis and specific findings, and then match figures by caption. We call this procedure citation-guided hierarchical image-pair annotation (CHIPA). On randomly selected case pairs, CHIPA finds about twice as many correct pairs as direct matching from full articles, and no incorrect ones. As the models read only text, the pairs are independent of any image model.

The dataset covers case reports published until February 2026, and we release the automated curation pipeline, which can be rerun as PMC grows. Each pair is annotated with the modality, anatomy and diagnosis its two images share, and we derived image-level labels in 12 modality, 20 anatomy and 40 diagnosis categories. We assigned all 86,893 images to training, validation and test partitions such that no pair or source article spans two partitions, preventing leakage between them. The release also includes citation contexts and generated citation-intent analyses, questions and answers for 121,831 case pairs.

Two experts reviewed a sample of pairs and judged the large majority to match and none to differ in diagnosis, and the image-level labels were also highly accurate. Fine-tuning the vision--language model BiomedCLIP\cite{biomedclip} on the pairs improves retrieval in every modality, anatomy and diagnosis category tested, both for queries consisting of an image alone (Fig.~2) and, more strongly, for queries that add a short text naming the finding of interest, as when a clinician searches for a specific abnormality. PMC-PaIR also supports representation learning, figure classification and, through the citation records and questions and answers, the study of citation intent and the training of models that compare clinical cases.

% ===========================================================================
\section{Methods}

Throughout, a \emph{case} is a patient record in PMC-Patients, and its \emph{source article} is the publication that reports it. Two cases form a \emph{case pair} when one source article cites the other; each case pair is counted once, regardless of citation direction. An \emph{image} is one published figure, which may contain several panels. An \emph{image pair} consists of one image from each source article of a case pair. The image pairs included in PMC-PaIR are \emph{positive pairs}, and the \emph{positives} of an image are the images that form a positive pair with it.

We selected citation-linked case pairs and their figures, annotated image pairs with CHIPA (Fig.~1), derived image-level labels and partitioned the images for benchmarking (Fig.~3). Table~1 gives the counts after each processing step and the model used at each stage of CHIPA.

\subsection{Selection of citation-linked case pairs}

We drew cases from two versions of PMC-Patients, the published collection of 167,034 cases from 140,897 articles published until January 2022\cite{pmcpatients,pmcpatients_data} and a collection of 319,939 cases from 268,151 articles published until February 2026, which we regenerated with the published construction code\cite{pmcpatients} from the PMC Open Access Subset and PubMed files downloaded in March 2026. We excluded records of the regenerated collection that lacked citation information. Because figures from an article that reports several patients cannot be reliably attributed to one of them, we retained only articles associated with exactly one case in their collection.

Candidate case pairs were the citation-based links that PMC-Patients records between the retained cases.\cite{pmcpatients} From the regenerated collection, we added only pairs containing at least one case absent from the published collection. Excluding cases without a candidate pair left 138,103 distinct cases in 159,388 candidate case pairs (Table~1).

\subsection{Figures, case descriptions and citation evidence}

For each selected case, we obtained the full-text Extensible Markup Language (XML) and figure files of the source article from the PMC Open Access Subset\cite{nlm_pmcoa_2026} and linked each figure and its caption to the case. To remove diagrams and text-heavy figures, we discarded images with predominantly white backgrounds and, among the remaining images, those in which Tesseract optical character recognition (through pytesseract) detected extensive text, using thresholds given in the code of the curation pipeline (Code Availability). We took the case description from the article sections with case-report headings such as `Case presentation' or `Case report'.

We verified each citation link by searching the reference list of each article for an entry identifying the other article, and a matching entry in either article confirmed the link. We matched entries on the digital object identifier (DOI), PubMed identifier (PMID) or PubMed Central identifier (PMCID), in that order, accepting a match on any one identifier and using normalized title substring matching as a fallback. As citation evidence, we collected the titles and abstracts of both articles and the citation contexts, that is, the paragraphs and table rows whose cross-reference (\texttt{<xref>}) elements point to the matched entry. For author--year citations that are not linked to the bibliography in the article XML, the link was confirmed from the reference list, but no citation context could be extracted.

We submitted a case pair for annotation only if the full-text XML and figure files of both articles were available, each case had at least one retained figure with a valid caption and a case description, and the citation link was verified. Of the 159,388 candidate case pairs, 122,365 met these criteria (Table~1).

\begin{table}[t]
\centering
\small
\caption{\textbf{Inputs, outputs and models of the annotation.} Counts are given after each processing step, and the exclusion rows give the case pairs removed by each selection criterion, applied in the order shown. All models are OpenAI models, and the reasoning-effort setting controls how much internal reasoning a model performs before answering.}

\begin{tabularx}{\linewidth}{@{}>{\raggedright\arraybackslash}Xr@{}}
\toprule
\textbf{Processing step} & \textbf{Count} \\
\midrule
Selected cases & 138,103 \\
Candidate case pairs & 159,388 \\
\hspace*{1em}Excluded: full-text XML or figure files unavailable & 2,669 \\
\hspace*{1em}Excluded: a case had no retained captioned figure & 31,969 \\
\hspace*{1em}Excluded: a case had no case description & 353 \\
\hspace*{1em}Excluded: citation link not confirmed & 2,032 \\
Case pairs submitted for annotation & 122,365 \\
Image pairs proposed & 93,113 \\
Case pairs removed by the veterinary-report filter & 534 \\
Case-pair citation records released & 121,831 \\
Image pairs released & 86,144 \\
\midrule
\textbf{Stage of CHIPA} & \textbf{Model (reasoning effort)} \\
\midrule
Stage 1 (citation-intent analysis) & \texttt{gpt-5.4-mini} (medium) \\
Stage 2 (question generation) & \texttt{gpt-5.4-nano} (high) \\
Stage 3 (question answering) & \texttt{gpt-5.4-nano} (high) \\
Stage 4 (image-pair proposal) & \texttt{gpt-5.4-mini} (low) \\
\bottomrule
\end{tabularx}
\end{table}

\subsection{Image-pair annotation with CHIPA}

CHIPA annotates each case pair with four successive LLM calls (Fig.~1): citation-intent analysis, question generation, question answering and image-pair proposal. Dividing the task gives each call a narrower objective and allows each stage to use a different model and reasoning-effort setting (Table~1). The prompts are reproduced at the end of Methods, and Table~2 shows the captions given to stage 4 and the outputs of stages 2--4 for one case pair.

\begin{table}[t]
\centering
\small
\caption{\textbf{Inputs and outputs of citation-guided hierarchical image-pair annotation (CHIPA) for one case pair.} Case~A\cite{pmc9899453} and case~B\cite{pmc4023003} describe children with paraspinal muscle infection complicated by spinal epidural abscess. One of the questions generated at each level is shown with its answer. Captions are paraphrased, and descriptors are reproduced from the model output. The image-level labels of \texttt{9899453\_1\_1} are derived from its three positive pairs, all of which have the descriptors shown. MRI, magnetic resonance imaging.}

\begin{tabularx}{\linewidth}{@{}>{\raggedright\arraybackslash}p{2.3cm}>{\raggedright\arraybackslash}p{3.2cm}>{\raggedright\arraybackslash}X@{}}
\toprule
\textbf{Stage} & \textbf{Item} & \textbf{Content} \\
\midrule
Stage 4 input & Caption of \texttt{9899453\_1\_1} (case~A) & T2-weighted and diffusion-weighted MRI findings of paraspinal and epidural abscesses, together with follow-up changes. \\
 & Caption of \texttt{4023003\_1\_3} (case~B) & Contrast-enhanced T1-weighted MRI showing paraspinal collections with thick enhancing walls and extension into the spinal canal. \\
\addlinespace
Stages 2 and 3 & Level 1: modality and organ system & Was spinal MRI the key test identifying paraspinal muscle infection with epidural extension in both cases? \emph{Yes. MRI identified paraspinal muscle infection with contiguous epidural abscess extension in both cases.} \\
 & Level 2: diagnostic category & Are both cases classified as pediatric paraspinal muscle infection complicated by spinal epidural abscess? \emph{Yes. Both describe a pediatric muscle abscess complicated by spinal epidural abscess.} \\
 & Level 3: specific findings & Did both cases show MRI findings supporting an abscess rather than diffuse cellulitis alone? \emph{Yes. Case~A shows diffusion-weighted epidural/posterior spinal canal findings consistent with an abscess. Case~B shows intramuscular fluid collections with thick enhancing walls.} \\
\addlinespace
Stage 4 output & \texttt{pair\_id} & \texttt{9899453\_1\_1}, \texttt{4023003\_1\_3} \\
 & \texttt{modality} & MRI \\
 & \texttt{anatomy} & lumbar spine/paraspinal muscles/epidural space \\
 & \texttt{diagnosis} & paraspinal abscess with spinal epidural abscess \\
\addlinespace
Image-level labels & Modality & \cat{mri} \\
 & Anatomy & \cat{musculoskeletal}, \cat{spine\_spinal\_cord} \\
 & Diagnosis & \cat{infection\_or\_abscess} \\
\bottomrule
\end{tabularx}
\end{table}

\parhead{Citation-intent analysis.} The model receives the titles and abstracts of both articles together with any extracted citation contexts. As in citation-intent classification,\cite{cohan_scaffolds_2019} it states the reason for the citation, together with the possible similarities between the two cases and a short explanation of how the citation is used. The purpose of a citation affects how informative the link is for comparing figures. For example, a report of an ectopic parathyroid adenoma in the piriform sinus\cite{pmc11366630} cites an earlier report of an adenoma at the same site,\cite{pmc10242159} whereas a report of an ectopic pituitary tumor\cite{pmc10641033} cites a report of metastatic testicular cancer\cite{pmc10065126} because both describe the use of ChatGPT in writing case reports. PMC-Patients, in contrast, treats every citation between two case reports as a link between similar patients, without examining its context.\cite{pmcpatients}

\parhead{Question generation and answering.} Given both case descriptions and the citation-intent analysis, the model generates about three comparative questions at each of three levels: imaging modality and organ system (level~1), diagnostic category and disease process (level~2), and specific imaging and clinical findings (level~3). A second call answers each question with Yes, No, Partial or Unclear and a brief reason, using only the case descriptions and the citation-intent analysis. The questions and answers provide the final stage with clinical context that captions may omit. The questions and answers for all 121,831 case pairs in the release are provided in \file{case_pair_questions_answers_eligible.jsonl} (Data Records).

\parhead{Image-pair proposal.} The final call receives the questions and answers together with the captions of both articles, each keyed by its image identifier, and considers every combination of one caption from each article. It returns, in JavaScript Object Notation (JSON), the combinations whose captions describe the same imaging modality and the same type of abnormality within the same anatomical region or organ system, together with free-text descriptors of the shared modality, anatomy and diagnosis. Differences in view, imaging plane, sequence, laterality, acuity, severity, slice or anatomical level do not by themselves prevent a match. The prompt restricts pairing to radiological, nuclear-medicine and ophthalmic imaging and excludes clinical and operative photographs, histopathology, endoscopy, electrophysiological recordings and diagrams. A case pair can yield zero, one or several image pairs.

\subsection{Post-processing and exclusion of veterinary reports}

Parsing the saved model responses, with formatting errors repaired where possible, yielded 93,113 proposed image pairs (Table~1). We then removed proposals with invalid image identifiers or unavailable image files, proposals involving source articles flagged as veterinary case reports (described below) and duplicate proposals (treating pairs as unordered), leaving 86,144 image pairs. Of the 121,831 case pairs that remained after the veterinary-report filter, 48,331 (39.7\%) contributed at least one positive pair.

To exclude veterinary case reports, we flagged likely dog and cat reports with keyword and phrase rules applied to article titles and abstracts, using journal titles as additional context. An article was flagged if its title matched an animal-related pattern, if its abstract matched such a pattern together with veterinary or case-report context, or if its abstract matched at least two distinct animal-related patterns. Exceptions covered expressions used in human case reports, such as `dog bite', `CAT scan' and the dental term `canine'. We applied the filter after annotation, removing 534 case pairs together with their image pairs.

\subsection{Image-level labels}

We derived image-level labels by mapping the free-text descriptors of the positive pairs to controlled vocabularies of 12 modality, 20 anatomy and 40 diagnosis categories, each of which includes the catch-all category \cat{other}. The 12 modality categories are CT, magnetic resonance imaging (MRI), radiography (including dental radiography and mammography), ultrasound, angiography/fluoroscopy, nuclear medicine, optical imaging, photography, microscopy, endoscopy, electrophysiology and \cat{other}. Optical imaging, assigned to 1,354 images, comprises optical coherence tomography and other ophthalmic or optical techniques such as fluorescein and indocyanine-green angiography, fundus autofluorescence, slit-lamp imaging and reflectance confocal microscopy, whereas fundus photographs are assigned to photography. Although the prompt restricts pairing to radiological, nuclear-medicine and ophthalmic imaging, some pairs share another modality, for example when a compound figure that combines CT with endoscopy is paired with an endoscopy image. Microscopy, endoscopy and electrophysiology labels are assigned to 1,449, 219 and 263 images, respectively, and photography labels to 837. The vocabularies and mapping rules are provided in \file{label_taxonomy.json}, and Table~2 shows an example. Regular-expression rules recognize spelling variants, synonyms and abbreviations. For example, `hepatic abscess' and `liver abscess' both map to \cat{infection\_or\_abscess}. Broad diagnostic categories group related terms: `tumor', `mass' and `unspecified lesion' all map to \cat{mass\_or\_lesion}, which does not distinguish benign from malignant findings.

The rules first split diagnosis descriptors into clauses at commas, semicolons and the connectives `but', `however', `and', `with', `plus', `complicated by', `associated with' and `due to'. Within a clause, rules for specific conditions take precedence over generic lesion terms, so that `abscess lesion' maps to \cat{infection\_or\_abscess}. The rules also recognize explicit negation and resolution, mapping, for example, `no abscess' to \cat{normal\_or\_negative\_report} and `resolved abscess' to \cat{postoperative\_or\_resolved\_state}. A descriptor that matches no rule maps to \cat{other}.

We assigned the categories derived from a pair to both of its images. For each image and field, we took the union of the categories contributed by all pairs that contain the image. An image can therefore carry several labels per field, and labels contributed by different pairs may refer to different panels of a compound figure. We assigned \cat{other} only when no pair contributed a more specific category to that field, which occurred for 65 images (0.07\%) in modality, 277 (0.32\%) in anatomy and 2,250 (2.59\%) in diagnosis.

The \cat{nuclear\_medicine} label is the only label also assigned directly from an image's own caption, independently of its pairs. Nuclear-medicine studies such as positron emission tomography (PET) are often fused with, or displayed beside, CT or MRI, whereas a pair descriptor records only the modality that the two images share. A fused PET/CT image paired with a CT image is therefore described as `CT'. We applied the \cat{nuclear\_medicine} rule of the taxonomy to every caption and assigned the label to each matching image, counting the abbreviation PET only when capitalized. Of the 3,068 images labeled \cat{nuclear\_medicine}, 873 received the label only from their captions.

\subsection{Benchmark partitions}

We partitioned the images so that every positive pair and every source article lies entirely within one partition (Fig.~3). We represented the dataset as an undirected graph whose nodes are images and whose edges are positive pairs. Each connected component of this graph contains the images linked directly or indirectly through positive pairs. Because the images of one source article can fall into different components, we merged components that share a source article until all images of each article belonged to the same merged component, which we call an \emph{article group}. Merging did not add pairs to the dataset and produced 13,338 article groups.

We sorted the article groups by their smallest image identifier, compared as strings, and shuffled them with the \texttt{shuffle} method of Python's \texttt{random.Random} generator initialized with seed 42. The first 10,670 groups (80\%, rounded down) formed the training partition, the next 1,333 (10\%, rounded down) the validation partition and the remaining 1,335 the test partition. These proportions apply to article groups rather than to images. The training, validation and test partitions contain 70,552, 8,245 and 8,096 images and 70,262, 8,034 and 7,848 positive pairs, respectively, and the test images come from 5,255 source articles.

\begin{figure}[t]
\centering
\includegraphics[width=\linewidth]{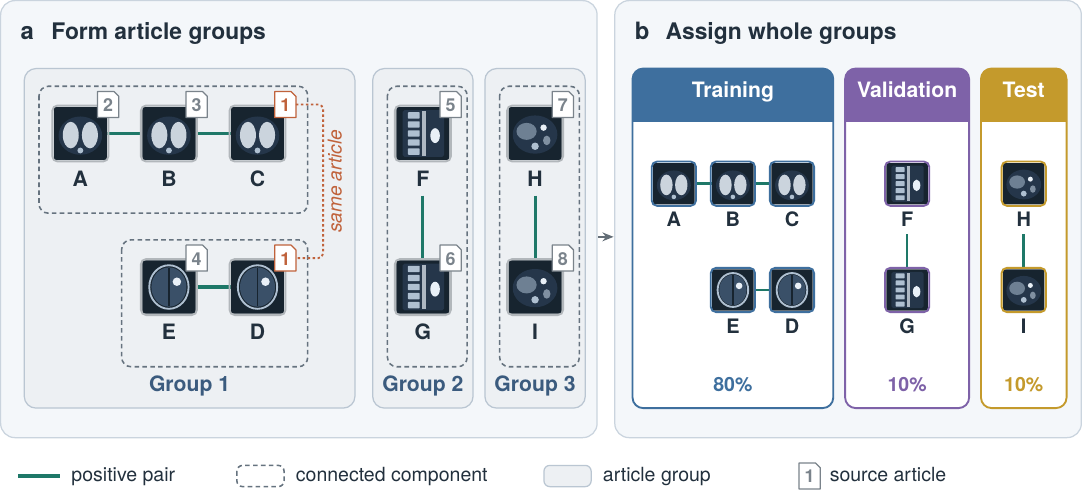}
\caption{\textbf{Partitioning by article groups.} \textbf{a},~Positive pairs link images, each tagged with the number of its source article, into connected components. Because images C and D come from the same source article (article~1) but lie in different components, their components are merged into one article group without adding a pair. \textbf{b},~Whole article groups are assigned to the training, validation and test partitions in proportions of 80:10:10 by number of groups, so that no positive pair or source article spans two partitions. Images are schematic.}

\end{figure}

\subsection{Ethics statement}

PMC-PaIR was derived entirely from case reports published in the PMC Open Access Subset and from the PMC-Patients collection extracted from them. No patients were recruited or contacted, no data other than the published articles were used, and no attempt was made to identify individuals.

\subsection{CHIPA prompts}

The prompts of the four CHIPA stages are reproduced verbatim from the curation pipeline. Text in braces is a placeholder that is replaced with the inputs of each case pair.

\paragraph{Stage 1: citation-intent analysis.} The placeholders are filled with the titles and abstracts of the citing and cited articles and with the extracted citation contexts.
\par\nolinenumbers
\begin{lstlisting}[style=llmprompt]
You are a medical citation analysis assistant. You will read excerpts from a 'citing paper' referencing another 'cited paper,' each describing a single-patient case report. Your task is to produce three pieces of information:

1. Reason for Citation (a short phrase or sentence about why the authors mention the cited paper).
2. Possible Similarities (a brief mention of how these patients’ cases align, if at all).
3. Explanation (one to two sentences summarizing how the citation is used).

----
Follow this exact output format:
Reason for Citation: <short phrase or sentence>
Possible Similarities: <brief mention>
Explanation: <one to two sentences>
----
Read the following excerpt(s) and generate your final answer in the required format. Unify your response if multiple citations are present. If no citations are specified, base your response on the articles' titles and abstracts.

Citing Title: {citing_title}
Citing Abstract: {citing_abstract}
Citation Paragraph: {citation_paragraph}
Cited Title: {cited_title}
Cited Abstract: {cited_abstract}
\end{lstlisting}
\linenumbers

\paragraph{Stage 2: question generation.} The placeholders are filled with the two case descriptions and the output of stage 1.
\par\nolinenumbers
\begin{lstlisting}[style=llmprompt]
You are an experienced clinician specializing in medical imaging. You will be given two cases and the reason for citation between the two (one case report cites the other), and asked to generate a set of comparative, clinically relevant questions that highlight both similarities and differences, helping differentiate potential diagnoses. Your questions should:

1. **Focus on Diagnostic Clues and Clinical Significance:**
   - Avoid trivial or purely technical details (like minor differences in image orientation).
   - Emphasize findings that inform diagnosis (e.g., presence of lesions, pattern of abnormalities, symptoms).

2. **Use a Comparative Framing:**
   - Explicitly compare Case A and Case B.
   - Aim for questions that probe both common ground and distinguishing features.
   - If possible, try asking mostly yes/no questions.

3. **Follow a Hierarchical Reasoning Approach:**
   - **Level 1 (Broad Context)**: Are both cases the same modality/organ system?
   - **Level 2 (General Diagnosis Category)**: Are they both infectious, both neoplastic, etc.?
   - **Level 3 (Specific Features & Findings)**: Detailed imaging findings (e.g., cavitation, consolidation, nodules), clinical presentation, organism type, etc.

4. **Consider Step-by-Step Reasoning:**
   - You may first silently analyze the diagnoses and imaging findings for each case.
   - Then generate questions that a clinician would naturally ask to tease out whether the two cases have the same underlying pathology or different pathologies.

5. **Provide a subtitle specifying the level (1/2/3) of the current set of questions.**

6. **Generate approximately three questions per hierarchical level.**

Below are the two case descriptions and the reason for citation between the two. Read them carefully, then generate a concise list of question prompts (in numbered format) that compare these cases on clinically meaningful aspects.

---
Case A:
{case_a}

Case B:
{case_b}

Reason for Citation, Possible Similarities and Explanation:
{citation_reason}
\end{lstlisting}
\linenumbers

\paragraph{Stage 3: question answering.} The placeholders are filled with the two case descriptions, the output of stage 1 and the questions from stage 2.
\par\nolinenumbers
\begin{lstlisting}[style=llmprompt]
You are an experienced clinician specializing in medical imaging.
You will be given:
- Case A
- Case B
- Reason for Citation
- A numbered list of comparative questions

Answer each question using only the provided case information.

Rules:
- Keep the exact numbering.
- Each line must follow this format exactly:
  <number>. <Yes|No|Partial|Unclear> — <brief reason>
- The brief reason must be short and evidence-based.
- Do not add any text before or after the numbered answers.
- Do not restate the questions.
- Do not use outside knowledge or speculation.

Below are the two case descriptions and the set of questions to be answered.

---
Case A:
{case_a}

Case B:
{case_b}

Reason for Citation, Possible Similarities and Explanation:
{citation_reason}

Questions:
{questions}
---
\end{lstlisting}
\linenumbers

\paragraph{Stage 4: image-pair proposal.} The placeholders are filled with the questions and answers from stages 2 and 3 and with the captions of the two articles, keyed by image identifier.
\par\nolinenumbers
\begin{lstlisting}[style=llmprompt]
You are a medical-AI expert.

You will receive:
1. Patient-context questions and answers for a case pair.
2. A dictionary of caption_id -> caption for Case A.
3. A dictionary of caption_id -> caption for Case B.

IMPORTANT PRE-FILTER: RADIOLOGY AND OPHTHALMIC IMAGING ONLY
Before forming any pairs, keep only captions that describe radiology imaging modalities or the ophthalmic imaging types listed below.
Allowed modality families and imaging types include:
- CT
- Ophthalmic photography / fundus imaging / OCT / OCTA / FA / ICG
- MRI
- X-ray / radiograph
- Fluoroscopy
- Mammography
- Ultrasound / sonography
- Angiography
- Nuclear medicine / hybrid imaging such as PET, SPECT, PET/CT, PET/MR

Ignore any caption outside these allowed radiology and ophthalmic imaging types, including:
- clinical photographs or gross inspection images, except the ophthalmic imaging listed above
- dermatology / operative / specimen photos
- pathology / histology / cytology / microscopy / IHC stains
- endoscopy / colonoscopy / bronchoscopy / EGD
- ECG / EKG / EEG
- diagrams / illustrations / schematics
- any other image outside the allowed radiology and ophthalmic imaging types

Your task is to evaluate every remaining inter-case caption pair (one caption from Case A and one caption from Case B) and return only the POSITIVE matches.

A pair is POSITIVE only if:
1. It is inter-case.
2. The two captions describe the same modality family, ignoring view/plane/sequence details.
3. The two captions describe the same pathology family within the same anatomy / organ system.
4. You ignore decorators such as acuity, severity, laterality, exact slice, or exact level.

Output rules:
- Output valid JSON only, with no markdown and no prose.
- Output a JSON array.
- Each item must have exactly these keys:
  {
    "pair_id": ["caption_id_from_case_a", "caption_id_from_case_b"],
    "modality": "<shared modality family>",
    "anatomy": "<shared anatomy or organ system>",
    "diagnosis": "<shared pathology family>"
  }
- Use caption IDs exactly as they appear in the provided dictionaries.
- If there are no positive matches, output [].

### PATIENT-CONTEXT QUERIES & ANSWERS
{questions}

{answers}

### IMAGE CAPTIONS (MULTIPLE PAIRS)
{captions_a}

{captions_b}
\end{lstlisting}
\linenumbers

% ===========================================================================
\section{Data Records}

PMC-PaIR is available from Zenodo.\cite{pmcpair_data} Table~3 lists the files. JSON Lines (JSONL) files contain one JSON object per line.

\parhead{Image and case identifiers.} Image identifiers link the annotation, label, partition, image and caption files. Each identifier joins, with underscores, the numeric part of the article's PMCID, the case index and the extraction index of the figure. For example, \texttt{9899453\_1\_1} is the first extracted figure of case~1 in article PMC9899453. Because only single-case articles were retained, the case index is 1 in every identifier. Each identifier refers to a complete figure, which may contain several panels. Case identifiers, which appear in the case-pair questions and answers, join the numeric part of the PMCID and the case index with a hyphen, as in case \texttt{9899453-1}, to which image \texttt{9899453\_1\_1} belongs.

\parhead{Image-pair annotations.} The file \file{image_pairs.json} is a JSON array with one object per positive pair. Each object has four fields: \texttt{pair\_id}, a two-element array with the identifiers of the two images, one from each source article; and the free-text descriptors \texttt{modality}, \texttt{anatomy} and \texttt{diagnosis}, which describe the shared imaging modality, anatomical region and type of abnormality. A descriptor may name several related structures or findings (Table~2). Each unordered pair occurs once.

\parhead{Image-level labels.} The file \file{image_labels.jsonl} contains one object per image with two fields, \texttt{image\_id} and \texttt{labels}, the second of which has the keys \texttt{modality}, \texttt{anatomy} and \texttt{diagnosis}, each holding an array of category names. The file \file{label_taxonomy.json} defines the categories and the mapping rules, including text normalization and matching priorities. Labels derived from pair descriptors can be traced to the records in \file{image_pairs.json} whose \texttt{pair\_id} contains the image identifier.

\parhead{Case-pair citation records.} The file \file{case_pair_citations_eligible.json} is a JSON array of 121,831 records, one for each annotated case pair that passed the veterinary-report filter. Each record describes the citation from one source article to the other with four string fields: \texttt{citing\_case} and \texttt{cited\_case}, the PMCIDs (for example, PMC9899453) of the citing and cited articles, whose numeric parts are the first components of the image identifiers; \texttt{extracted\_citation}, the citation contexts; and \texttt{citation\_reason}, the output of the citation-intent analysis stage. In the 857 records for which no citation context could be extracted, \texttt{extracted\_citation} is an empty string.

\parhead{Case-pair questions and answers.} The file \file{case_pair_questions_answers_eligible.jsonl} contains 121,831 records, one for each case pair in the citation records. Each record has eight string fields: \texttt{annotation\_id}, a unique record identifier; \texttt{case\_pair\_id}, an order-independent identifier that joins the two case identifiers, sorted as strings, with `|', for example \texttt{5563556-1|8167975-1}; \texttt{case\_a\_uid} and \texttt{case\_b\_uid}, the identifiers of the two cases in the order used in the prompts (case~A and case~B), which does not indicate citation direction; \texttt{citing\_case} and \texttt{cited\_case}, the PMCIDs of the citing and cited articles, for example PMC8167975 and PMC5563556; and \texttt{questions} and \texttt{answers}, the generated text as saved from the model output. Records can be joined with the citation records through \texttt{citing\_case} and \texttt{cited\_case}.

\parhead{Images and captions.} The archives \file{images_captions_001.zip} to \file{images_captions_004.zip} contain 86,825 images, stored as \file{images/<image_id>.jpg}, and their source captions, stored as \file{captions/<image_id>.txt}. Files for the remaining 68 images are withheld because the redistribution terms of their article or figure could not be confirmed (44 images), because their licenses prohibit derivatives (23) or because they await privacy review (1). Of these images, 60 are in the training partition, 3 in the validation partition and 5 in the test partition, and \file{withheld.csv} lists their identifiers, the reasons and their locations in the source articles. All 86,893 identifiers are retained in the annotation and label files, and \file{pair_availability.csv} indicates for every positive pair whether the files of both images are distributed. The file \file{file_manifest.csv} gives the archive, size and SHA-256 checksum of every image and caption file.

\parhead{Benchmark partitions.} The partition of every image is distributed with the code, as \file{data/article_disjoint_split.json} in the retrieval repository and, as an identical copy, as \file{image_splits.json} in the dataset-curation repository (Code Availability).

\parhead{Licensing and provenance.} The file \file{licenses.csv} records the provenance, publication year, license and attribution information of the source articles and of the distributed images and captions. Images, captions and the citation contexts in the case-pair records remain under the terms of their source articles. Our annotations, labels, generated citation-intent analyses, questions and answers, and other original content are released under the Creative Commons Attribution 4.0 International license (CC BY 4.0) to the extent of the authors' rights, as stated in \file{LICENSE.txt}.

\begin{table}[t]
\centering
\small
\caption{\textbf{Files in the PMC-PaIR data record.}}

\begin{tabularx}{\linewidth}{@{}>{\raggedright\arraybackslash}p{5.6cm}l>{\raggedright\arraybackslash}X@{}}
\toprule
\textbf{File} & \textbf{Format} & \textbf{Contents} \\
\midrule
\file{image_pairs.json} & JSON & 86,144 positive pairs with shared modality, anatomy and diagnosis descriptors \\
\file{image_labels.jsonl} & JSONL & Modality, anatomy and diagnosis labels of 86,893 images \\
\file{label_taxonomy.json} & JSON & Label categories and mapping rules \\
\file{case_pair_citations_eligible.json} & JSON & Citation contexts and citation-intent analyses of 121,831 case pairs \\
\file{case_pair_questions_answers_eligible.jsonl} & JSONL & Generated questions and answers for 121,831 case pairs \\
\file{images_captions_001.zip} to \file{images_captions_004.zip} & ZIP & 86,825 images (JPEG) and their captions (text) \\
\file{file_manifest.csv} & CSV & Archive, size and SHA-256 checksum of each image and caption file \\
\file{licenses.csv} & CSV & Provenance, publication year, license and attribution \\
\file{withheld.csv} & CSV & Identifiers, reasons and source locations of the 68 withheld images \\
\file{included_image_ids.txt} & Text & Identifiers of the 86,825 distributed images \\
\file{withheld_image_ids.txt} & Text & Identifiers of the 68 withheld images \\
\file{pair_availability.csv} & CSV & Availability of the image files of both members of each positive pair \\
\file{release_summary.json} & JSON & Release counts and archive and licensing summaries \\
\file{README.txt}, \file{LICENSE.txt}, \file{SHA256SUMS.txt} & Text & File descriptions and reuse instructions, licensing statement and file checksums \\
\bottomrule
\end{tabularx}
\end{table}

\parhead{Dataset composition.} The 86,144 positive pairs involve 86,893 images from 56,013 source articles. The articles were published between 1986 and 2026, 60.6\% of them in 2020 or later, with a median year of 2021 and an interquartile range (IQR) of 2017--2023 (Table~4). CT, MRI and radiography are the most frequent modality labels (46.4\%, 23.1\% and 15.5\% of images), and the most frequent anatomy and diagnosis labels are thorax/respiratory (14.8\%) and mass/lesion (31.4\%), respectively. An image belongs to a median of one positive pair (IQR 1--2; maximum 33), and 45.9\% of images belong to two or more. The graph of positive pairs has 18,597 connected components with a median of three images (IQR 2--4; maximum 963), and the 13,338 article groups contain a median of three images (IQR 2--6; maximum 1,465).

\begin{table}[t]
\centering
\small
\caption{\textbf{Composition of PMC-PaIR.} Percentages for labels and for positive pairs per image are relative to all 86,893 images, including the 68 whose files are withheld, and those for publication years to the 56,013 source articles. Label rows show the three most frequent categories of each field, with display names (for example, Mass/lesion for \cat{mass\_or\_lesion}). An image can carry several labels per field.}

\begin{tabular}{@{}llr@{}}
\toprule
\textbf{Measure} & \textbf{Category} & \textbf{Count (\%)} \\
\midrule
Modality labels & CT & 40,327 (46.4) \\
 & MRI & 20,104 (23.1) \\
 & Radiography & 13,459 (15.5) \\
\addlinespace
Anatomy labels & Thorax/respiratory & 12,851 (14.8) \\
 & Musculoskeletal & 12,127 (14.0) \\
 & Vasculature & 11,827 (13.6) \\
\addlinespace
Diagnosis labels & Mass/lesion & 27,277 (31.4) \\
 & Infection/abscess & 8,361 (9.6) \\
 & Inflammatory/autoimmune & 6,623 (7.6) \\
\addlinespace
Positive pairs per image & 1 & 47,002 (54.1) \\
 & 2 & 20,173 (23.2) \\
 & 3--5 & 16,364 (18.8) \\
 & $\geq$6 & 3,354 (3.9) \\
\addlinespace
Publication year of source article & 1986--1999 & 18 ($<$0.1) \\
 & 2000--2009 & 1,443 (2.6) \\
 & 2010--2019 & 20,636 (36.8) \\
 & 2020--2026 & 33,916 (60.6) \\
\bottomrule
\end{tabular}
\end{table}

% ===========================================================================
\section{Technical Validation}

We assessed the citation-intent analyses that guide annotation, the structured design of CHIPA compared with single-step annotation, the image pairs, the image-level labels and the use of the pairs for training and evaluating image retrieval. Table~5 summarizes the manual checks. For proportions from manual review, we report exact (Clopper--Pearson) 95\% confidence intervals (CIs).

\begin{table}[t]
\centering
\small
\caption{\textbf{Summary of manual validation.} Percentages are followed by exact 95\% confidence intervals (CIs). Relative recall is the number of correct pairs a workflow proposed divided by the number of correct pairs proposed by either workflow. Radiologist-rated Hit@5 is the proportion of queries for which at least one of the five highest-ranked candidates matched the query in the stated attribute or attributes. (i+t)2i, image-plus-text retrieval.}

\begin{tabularx}{\linewidth}{@{}>{\raggedright\arraybackslash}p{3.0cm}>{\raggedright\arraybackslash}p{3.9cm}>{\raggedright\arraybackslash}X@{}}
\toprule
\textbf{Check} & \textbf{Sample} & \textbf{Outcome} \\
\midrule
Citation-intent analyses & 25 randomly sampled case pairs & Supported 22 (88.0\%; 68.8--97.5); partly supported 2 (8.0\%; 1.0--26.0); not supported 1 (4.0\%; 0.1--20.4) \\
\addlinespace
CHIPA versus single-step annotation & 50 randomly sampled case pairs; all 29 image pairs proposed by either workflow & CHIPA: precision 25/25 (100.0\%; 86.3--100.0); relative recall 25/27 (92.6\%; 75.7--99.1). Single-step: precision 12/14 (85.7\%; 57.2--98.2); relative recall 12/27 (44.4\%; 25.5--64.7) \\
\addlinespace
Image pairs & 163 randomly sampled pairs; two independent reviewers, then consensus & Yes 147 (90.2\%; 84.5--94.3); Borderline 6 (3.7\%; 1.4--7.8); No 10 (6.1\%; 3.0--11.0); unweighted Cohen's $\kappa$ = 0.70 before consensus \\
\addlinespace
Image-level labels & 320 images in the reviewed pairs & Yes 299 (93.4\%; 90.1--95.9); Borderline 1 (0.3\%; 0.0--1.7); No 20 (6.3\%; 3.9--9.5) \\
\addlinespace
Top five candidates of the fine-tuned (i+t)2i model & 17 randomly sampled CT test queries with heart or vasculature labels; 85 query--candidate pairs; one board-certified radiologist & Radiologist-rated Hit@5: modality 17/17 (100.0\%; 80.5--100.0); anatomy 17/17 (100.0\%; 80.5--100.0); diagnosis 12/17 (70.6\%; 44.0--89.7); all three 11/17 (64.7\%; 38.3--85.8) \\
\bottomrule
\end{tabularx}
\end{table}

\subsection{Citation-intent analyses}

To check that the citation-intent analyses are faithful to their inputs, we sampled 25 case pairs at random and compared each analysis with the titles, abstracts and citation contexts from which it was generated. An analysis was rated \emph{supported} when it correctly described the role of the cited article, \emph{partly supported} when it identified relevant content but overstated or imprecisely described that role, and \emph{not supported} when the source text did not establish the claimed reason. Twenty-two analyses were supported, two were partly supported and one was not supported (Table~5).

\subsection{CHIPA versus single-step annotation}

We compared CHIPA with a single-step baseline that received the full-text XML of both articles and proposed image pairs directly, without citation-intent analysis or questions and answers, using the model and reasoning-effort setting of the image-pair proposal stage (\texttt{gpt-5.4-mini}, low). Both workflows were applied to 50 randomly selected case pairs, and every image pair proposed by either was reviewed manually. Of the 29 distinct pairs proposed, 27 were judged correct. CHIPA proposed 25 of these 27 pairs and no incorrect pair, whereas the baseline proposed 12 correct and 2 incorrect pairs (Table~5). Using \texttt{o4-mini} (low reasoning effort) in place of \texttt{gpt-5.4-mini} gave similar results. The comparison focused on proposed pairs because PMC-PaIR lists positive pairs only. A false positive adds an incorrect pair to the dataset, whereas a missed pair leaves a true pair unlisted.

\subsection{Expert review of image pairs}

A medical physicist and a resident physician independently reviewed 163 randomly sampled pairs and resolved disagreements by discussion. For each pair, the reviewers examined both images, the titles and abstracts of the two source articles and the citation context, and rated the match in modality, anatomy and diagnosis as Yes, Borderline or No. The overall rating was No if any attribute was rated No and otherwise equaled the diagnosis rating.

By consensus, 147 pairs (90.2\%), including 13 with a borderline modality or anatomy rating, matched overall, 6 (3.7\%) were borderline and 10 (6.1\%) did not match (Fig.~4a and Table~5). The 10 mismatches concerned modality (4 pairs) or anatomy (6 pairs), and diagnosis, which matched for 156 pairs (95.7\%), was never rated a mismatch. Before discussion, the reviewers' overall ratings agreed for 151 pairs (92.6\%; 95\% CI 87.5--96.1\%), with an unweighted Cohen's $\kappa$ of 0.70, which indicates substantial agreement.\cite{landis1977}

\begin{figure}[t]
\centering
\includegraphics[width=\linewidth]{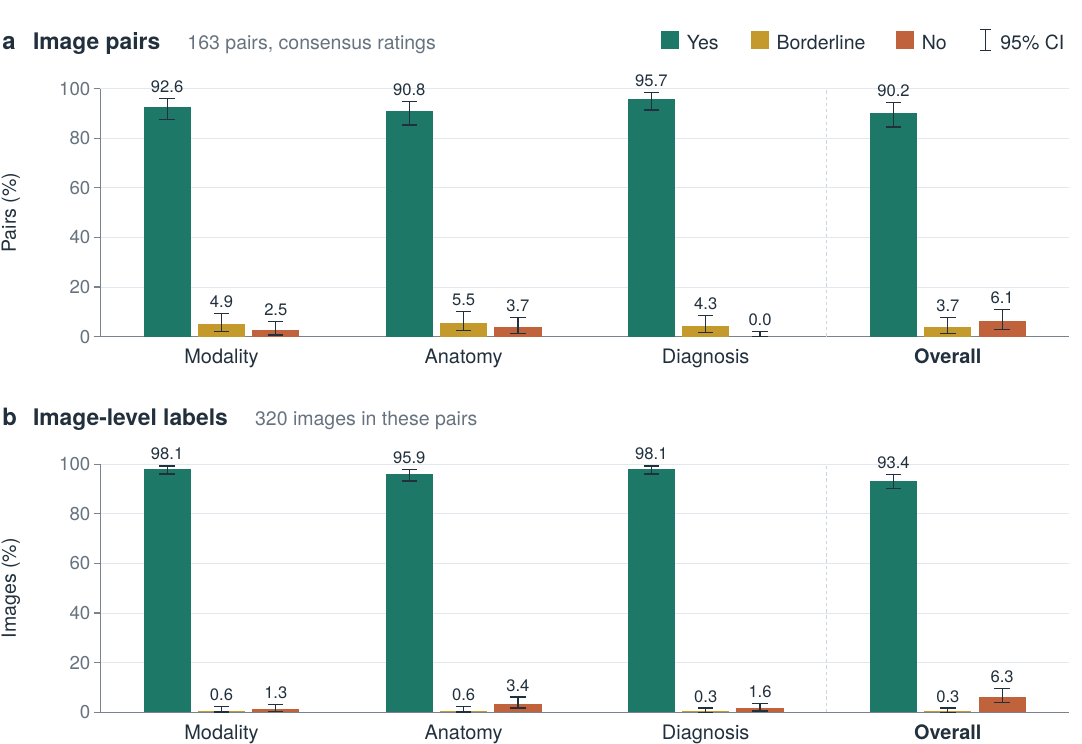}
\caption{\textbf{Manual review of image pairs and image-level labels.} \textbf{a},~Consensus ratings of a medical physicist and a resident physician for 163 randomly sampled pairs. Modality, anatomy and diagnosis were rated separately, and the overall rating is No if any attribute is rated No and otherwise equals the diagnosis rating. \textbf{b},~Ratings of the labels of the 320 images in these pairs, with the overall rating derived in the same way. Bars and numbers give the percentage of pairs or images with each rating, and error bars give exact (Clopper--Pearson) 95\% confidence intervals, computed separately for each rating.}

\end{figure}

\subsection{Review of image-level labels}

The labels of the 320 distinct images in the 163 reviewed pairs were also rated. For each field, labels were rated Yes when supported by the reviewed evidence, Borderline when they described only part of the main abnormality and No when a category was incorrect. A label was counted as supported if the evidence of any pair containing the image supported it. The overall rating followed the rule used for pairs. Labels were correct overall for 299 images (93.4\%), borderline for 1 (0.3\%) and incorrect for 20 (6.3\%; Fig.~4b and Table~5). Modality, anatomy and diagnosis labels were correct for 314 (98.1\%), 307 (95.9\%) and 314 (98.1\%) images, respectively. Of the 20 errors, 12 arose when descriptors were mapped to categories, 7 were inherited from the pair descriptors and 1 had both sources.

\subsection{Image retrieval with the benchmark partitions}

\parhead{Benchmark design.} To test whether the positive pairs can be used to train and evaluate medical image retrieval models, we fine-tune a vision--language model on the training partition and evaluate it on the test partition. We use BiomedCLIP, which was pretrained on the 15 million PMC figure--caption pairs of PMC-15M.\cite{biomedclip} Each of the 8,096 test images serves once as a query. Its candidates are all other test images except those linked to it only indirectly, through other positive pairs in its connected component, which we exclude because they may be clinically related to the query. To cover wide and multi-panel figures, we resize each image so that its shorter side is 224 pixels, take three $224 \times 224$ crops from the start, middle and end of its longer side, and represent the image by the normalized mean of the three crop embeddings.

We evaluate two retrieval tasks (Fig.~1). In image-only retrieval (i2i), candidates are ranked by the cosine similarity between the query and candidate image embeddings. In image-plus-text retrieval, abbreviated (i+t)2i, the query image is accompanied by the descriptions of its positive pairs, which are sentences generated from the pair descriptors with the template `\{diagnosis\} in the \{anatomy\} on \{modality\}.' Candidates are ranked by the unweighted sum of the image similarity and the mean similarity between the candidate image and the query's distinct descriptions. This task evaluates retrieval when a description of the finding of interest, here derived from the pair annotations, is available at query time. We compare each fine-tuned model with pretrained BiomedCLIP queried in the same way.

\parhead{Evaluation.} Hit@$k$ is the proportion of queries with at least one positive among the $k$ highest-ranked candidates, Recall@$k$ is the mean fraction of each query's positives found among them, and the mean reciprocal rank (MRR) is the mean of the reciprocal rank of the first positive,\cite{manning_informationretrieval_2008} with each query weighted equally. We estimate 95\% CIs for the differences between fine-tuned and pretrained models with a cluster bootstrap\cite{field_bootstrap_2007} that resamples whole test article groups 5,000 times, keeping all queries of each drawn group together and evaluating both models on the same resampled queries. Because relevance is defined by the positive pairs, the metrics measure how well a model retrieves the annotated positives rather than all clinically similar images in the test partition.\cite{buckley_incomplete_2004}

\parhead{Fine-tuned models.} In both tasks, the eligible negatives of a training image are the images not linked to it, directly or indirectly, through positive pairs, and we select the checkpoint with the highest validation Recall@10, computed with the same candidates, image crops and ranking as at test time. For i2i, we fine-tune all 12 transformer blocks of the image encoder with layer-wise learning-rate decay and an InfoNCE contrastive loss,\cite{oord_cpc_2018,radford_clip_2021} pairing each training image in each epoch with one randomly chosen positive and using the other eligible images in the batch as negatives. For (i+t)2i, we fine-tune the text encoder and all but the first eight blocks of the image encoder with two triplet-ranking losses:\cite{faghri_vsepp_2018} an image--image loss and a bidirectional image--text loss that aligns each pair description with the normalized mean embedding of the pair's two images. Training batches for this model include hard negatives,\cite{robinson_hardnegatives_2021} that is, images that pretrained BiomedCLIP rates as similar to a training image but that belong to a different connected component and a different source article. Table~6 lists all settings.

\begin{table}[t]
\centering
\small
\caption{\textbf{Fine-tuning settings for the two retrieval tasks.} i2i, image-only retrieval; (i+t)2i, image-plus-text retrieval. Both models use three $224 \times 224$ crops per image, the AdamW optimizer, bfloat16 mixed precision, gradient clipping at 1.5, and 200 warm-up steps followed by cosine learning-rate decay scheduled over the maximum number of epochs. Checkpoints were selected by validation Recall@10.}

\begin{tabularx}{\linewidth}{@{}>{\raggedright\arraybackslash}p{3.4cm}>{\raggedright\arraybackslash}X>{\raggedright\arraybackslash}X@{}}
\toprule
\textbf{Setting} & \textbf{i2i} & \textbf{(i+t)2i} \\
\midrule
Text input & None & Pair descriptions of up to 256 tokens \\
Trained parameters & Image encoder, with layer-wise learning-rate decay of 0.75 & Text encoder and image encoder except blocks 1--8 \\
Loss & InfoNCE, temperature 0.10 & Image--image (weight 0.3) and image--text (weight 0.6) triplet-ranking losses, margin 0.09, over the mean score of up to four highest-scoring eligible negatives \\
Negatives & Eligible images in the batch & Images and descriptions in the batch; batches assembled with mined hard negatives \\
Augmentation & Crop jitter, brightness and contrast jitter, rotation of up to 5$^\circ$ & Crop jitter \\
Batch size & 128 anchor--positive pairs & 128 image pairs \\
Peak learning rate; weight decay & $6 \times 10^{-6}$; 0.05 & $1 \times 10^{-6}$; 0.2 \\
Epochs & 12 & At most 30, with early stopping after three epochs without improvement \\
Selected checkpoint (seed 42) & Epoch 11 & Epoch 21 (training stopped after epoch 24) \\
\bottomrule
\end{tabularx}
\end{table}

\begin{table}[!t]
\centering
\small
\caption{\textbf{Retrieval performance on the 8,096 test queries.} i2i, image-only queries; (i+t)2i, image-plus-text queries. Hit@10 is the proportion of queries with at least one positive among the ten highest-ranked candidates, Recall@10 the mean fraction of positives found among them, MRR the mean reciprocal rank of the first positive and median rank the median rank of the first positive. Each fine-tuned model (training seed 42) is compared with pretrained BiomedCLIP queried in the same way. Differences are fine-tuned minus pretrained values, computed before rounding, with 95\% confidence intervals (CIs) from a cluster bootstrap over test article groups.}

\begin{tabular}{@{}llrrr@{}}
\toprule
\textbf{Configuration} & \textbf{Metric} & \textbf{Pretrained} & \textbf{Fine-tuned} & \textbf{Difference (95\% CI)} \\
\midrule
i2i & Recall@10 (\%) & 14.62 & 20.15 & 5.53 (4.61 to 6.55) \\
 & Hit@10 (\%) & 21.47 & 28.31 & 6.84 (5.73 to 8.09) \\
 & MRR & 0.111 & 0.143 & 0.033 (0.027 to 0.039) \\
 & Median rank & 65 & 37 & \\
\addlinespace
(i+t)2i & Recall@10 (\%) & 20.53 & 31.62 & 11.09 (9.89 to 12.38) \\
 & Hit@10 (\%) & 29.68 & 42.98 & 13.30 (12.07 to 14.66) \\
 & MRR & 0.154 & 0.220 & 0.065 (0.057 to 0.075) \\
 & Median rank & 34 & 15 & \\
\bottomrule
\end{tabular}
\end{table}

\FloatBarrier
\begin{figure}[p]
\centering
\includegraphics[width=\linewidth]{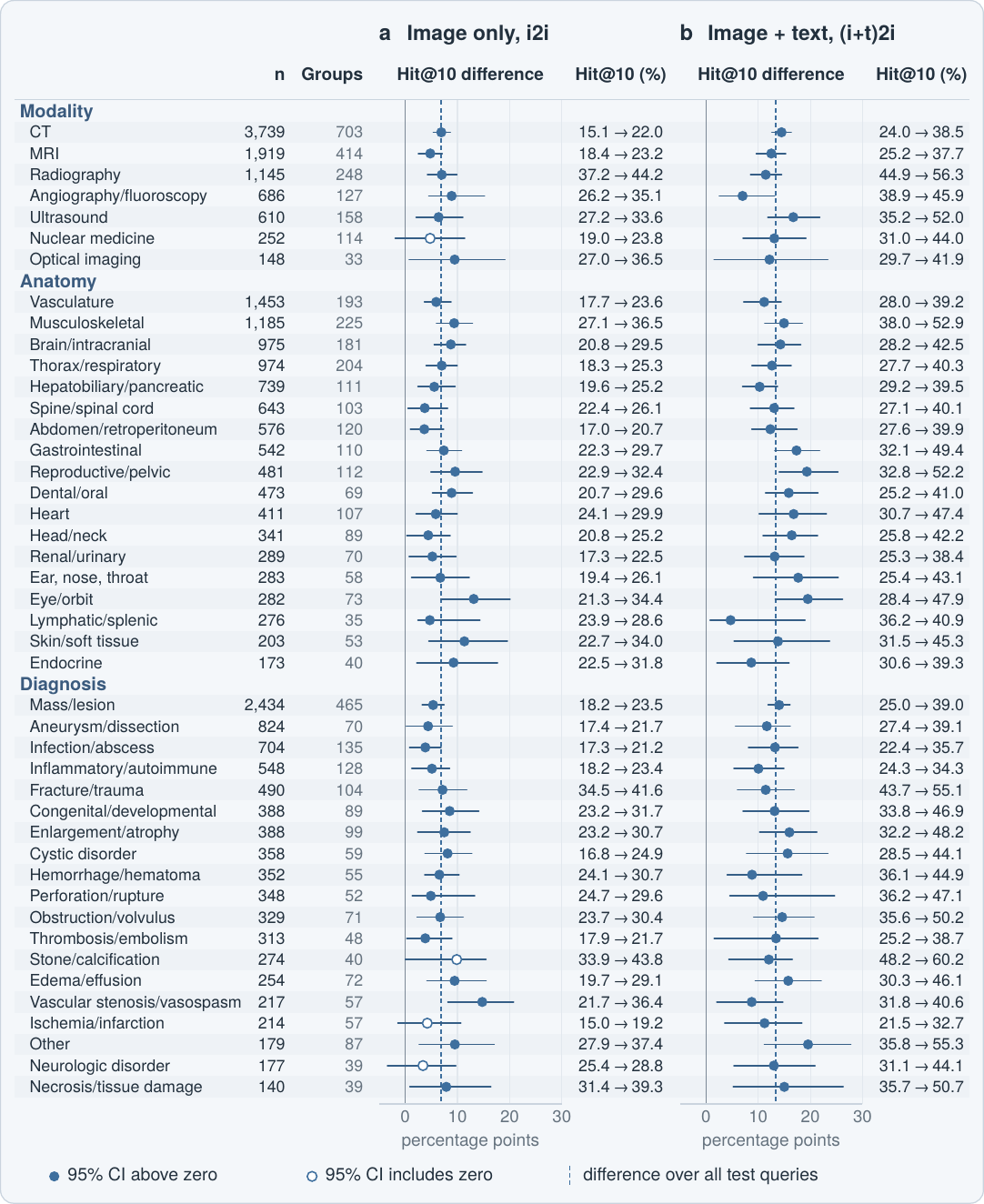}
\caption{\textbf{Change in Hit@10 with fine-tuning across label categories of the test queries.} Hit@10 is the proportion of queries with at least one positive among the ten highest-ranked candidates. \textbf{a},~Image-only queries, i2i. \textbf{b},~Image-plus-text queries, (i+t)2i. For the queries that carry each label, points show the fine-tuned minus pretrained Hit@10, in percentage points, with 95\% confidence intervals from 5,000 paired cluster-bootstrap resamples of article groups. The difference over all test queries is 6.84 percentage points for i2i and 13.30 for (i+t)2i. Columns on the left give the number of queries ($n$) and article groups in each category, and numbers on the right give Hit@10 (\%) of the pretrained and fine-tuned models. Categories with at least 140 queries from at least 20 article groups are shown. A query can carry several labels per field, and intervals are not adjusted for multiple comparisons.}

\end{figure}

\parhead{Retrieval results.} Fine-tuning improves all metrics in both tasks (Table~7). The gains are larger when the query includes text, with Recall@10 rising by 11.09 points for (i+t)2i and by 5.53 points for i2i. For i2i, Hit@10 increases from 21.5\% to 28.3\% and the median rank of the first positive decreases from 65 to 37; for (i+t)2i, Hit@10 increases from 29.7\% to 43.0\%. Across training seeds 42, 43 and 44, test Recall@10 is $20.11 \pm 0.07\%$ for i2i and $31.57 \pm 0.04\%$ for (i+t)2i (mean $\pm$ s.d.). Among the 44 modality, anatomy and diagnosis categories with at least 140 test queries from at least 20 article groups, Hit@10 increases in every category in both tasks, and the 95\% CI excludes zero in 40 categories for i2i and in all 44 for (i+t)2i (Fig.~5). Figure~2 shows an example query.

\subsection{Radiologist review of retrieved images}

A board-certified radiologist with 9 years of experience (K.K.B.) reviewed the five highest-ranked candidates returned by the fine-tuned (i+t)2i model for 17 test queries (85 query--candidate pairs). The queries were sampled at random from those with CT and heart or vasculature labels, to match the radiologist's subspecialty. The radiologist rated whether each candidate matched the query in modality, anatomy and diagnosis, regardless of whether it was one of the query's positives. At least one of the five candidates matched the query in modality and in anatomy for all 17 queries, in diagnosis for 12 (70.6\%) and in all three attributes for 11 (64.7\%; Table~5). These ratings count any clinically matching candidate and therefore complement the metrics based on the positive pairs.

% ===========================================================================
\section{Usage Notes}

\parhead{Benchmarking.} Because the pairs were derived from text, without reference to any image model, they can be used to compare image models. To obtain results comparable with those reported here, users should adopt the released partitions and, as in Technical Validation, exclude from each query's candidates the images linked to it only indirectly. The reported results use all 8,096 test images, including 5 whose files are withheld, and can be recomputed from the saved per-query results (Code Availability). With the distributed files, the retrieval code excludes the withheld images and their pairs and evaluates the remaining 8,088 test queries, and the resulting metrics differ slightly from those reported. Models pretrained on PMC figures and captions, including BiomedCLIP,\cite{biomedclip} may have encountered PMC-PaIR images during pretraining.

\parhead{Positive pairs.} PMC-PaIR contains positive pairs only; image pairs that are not listed are unlabeled rather than negative.\cite{elkan_positive_2008} When training with contrastive objectives, images linked only indirectly through positive pairs can be excluded from the negatives, as in our fine-tuning experiments, or false negatives can be handled explicitly.\cite{wang_medclip_2022} By design, the two images of a pair share an imaging modality, which excludes cross-modality matches such as a CT and an MRI image of the same kind of lesion.

\parhead{Labels and descriptors.} Shared image-level labels indicate a coarser similarity than positive pairs: two images labeled CT, thorax/respiratory and mass/lesion can show unrelated abnormalities, whereas the two images of a positive pair were matched on the type of abnormality in the context of both cases. Because labels describe whole figures, which may contain several panels, users who need panel-level labels can combine PMC-PaIR with compound-figure separation.\cite{subramanian_medicat_2020} Users who need finer or different categories can remap the free-text descriptors with their own rules or adapt those in \file{label_taxonomy.json}, and label errors that arise in this mapping can be corrected in the same way, without changing the pair annotations. To restrict the data to radiological images, users can filter images and their pairs by modality label.

\parhead{Images and licensing.} Images are figures as published in the source articles and are distributed in JPEG format. Many contain arrows, letters or other marks added by the authors, which users training image models may want to crop or mask. The images retain the licenses of their source articles, and \file{licenses.csv} allows users to select images by license and to attribute them.

\parhead{Linking to the source articles.} The PMCID numbers in the image identifiers and the PMCIDs in the citation records link every image to its source article in PMC, from which users can retrieve the full case description and the article's other figures, for example to build case-level or combined text-and-image tasks.

\parhead{Questions and answers.} For each case pair, the released questions and answers record the clinical comparison that guided image matching, at the levels of imaging modality and organ system, diagnosis and specific findings. They can explain why two images were paired, and their answers at each level allow users to select case pairs that agree at a chosen level, for example to build stricter or graded relevance judgments. They can also serve as training and evaluation data for models that compare two clinical cases or generate comparative clinical questions, and as intermediate supervision for distilling the annotation procedure into smaller open models.

\parhead{Extending the dataset.} The released curation pipeline can be rerun on later releases of the PMC Open Access Subset to annotate newly published case reports, given application programming interface (API) access to language models such as those in Table~1. Unlike the procedure described in Methods, it applies the veterinary-report filter before annotation.

% ===========================================================================
\section{Data Availability}

PMC-PaIR is available from Zenodo\cite{pmcpair_data} at \url{https://doi.org/10.5281/zenodo.23439169}. The record contains the image-pair annotations, image-level labels, label taxonomy, case-pair citation records, case-pair questions and answers, 86,825 images with their captions, and the provenance, licensing and checksum files described in Data Records. The benchmark partitions are distributed with the code, which is archived at Zenodo.

\section{Code Availability}

Code for the dataset-curation pipeline is available in the \href{https://github.com/GuyShmueli/pmc-pair-pipeline}{\texttt{pmc-pair-pipeline} GitHub repository}, together with the annotation prompts and the provenance records of the dataset construction. Code for BiomedCLIP fine-tuning, retrieval evaluation and analysis of the saved results is available in the \href{https://github.com/GuyShmueli/pmc-pair-retrieval}{\texttt{pmc-pair-retrieval} GitHub repository}, together with the benchmark partitions, the saved evaluation results, which include per-query results and overall and per-category metrics (\file{results/figures/source_data/subgroups.csv}), and the metadata and checksums of the fine-tuned checkpoints. The fine-tuned weights are attached to release v1.0.3 of this repository under the MIT License, together with the license notice of BiomedCLIP. Version 1.0.2 of the pipeline repository and version 1.0.3 of the retrieval repository are archived at Zenodo.\cite{shmueli_pipeline_2026,shmueli_retrieval_2026} They differ from version 1.0.0, which was used for this work, only in licensing, documentation, checksums, version strings and, for the retrieval code, the download of the fine-tuned weights from GitHub. Both repositories include environment specifications, configuration files and usage instructions. Their code, documentation, configuration files and prompts are released under the MIT License, and their data files, including the benchmark partitions and the label taxonomy, and saved results under CC BY 4.0. Article images reproduced in the example figures of the repositories keep the licenses of their source articles. Fine-tuning started from the pretrained checkpoint \path{microsoft/BiomedCLIP-PubMedBERT_256-vit_base_patch16_224}. The LLMs and reasoning-effort settings are listed in Table~1.

% ===========================================================================
\begingroup
\small
\setlength{\parskip}{0pt}
\renewcommand{\refname}{References}

\endgroup

% ===========================================================================
\section{Author Contributions}

G.S. designed and implemented the annotation workflow, built the dataset and benchmark, performed the technical validation and wrote the manuscript. K.K.B. reviewed the retrieved images and revised the manuscript. T.H. conceived and supervised the project and revised the manuscript.

\section{Competing Interests}

The authors declare no competing interests.

\section{Funding}

This work received no external funding.

\end{document}